\pdfoutput=1
\documentclass[11pt]{article}

\usepackage[]{ACL2023}

\usepackage{times}
\usepackage{latexsym}

\usepackage[T1]{fontenc}
\usepackage[utf8]{inputenc}

\usepackage{microtype}

\usepackage{inconsolata}
\usepackage{booktabs}
\usepackage{amsfonts,amsmath}

\title{BM25 Query Augmentation Learned End-to-End}

\author{Xiaoyin Chen \and Sam Wiseman \\
        Duke University \\
        \texttt{\{xc177, swiseman\}@cs.duke.edu}}

\newcommand{\boldv}{\mathbf{v}}

\newcommand{\boldu}{\mathbf{u}}

\newcommand{\boldh}{\mathbf{h}}
\newcommand{\boldc}{\mathbf{c}}

\newcommand{\bolda}{\mathbf{a}}

\newcommand{\boldf}{\mathbf{f}}
\newcommand{\boldw}{\mathbf{w}}

\newcommand{\boldW}{\mathbf{W}}

\newcommand{\mcV}{\mathcal{V}}

\newcommand{\mcL}{\mathcal{L}}
\newcommand{\mcD}{\mathcal{D}}

\newcommand{\reals}{\mathbb{R}}

\newcommand{\enc}{\mathbf{enc}}

\DeclareMathOperator*{\RELU}{ReLU}

\begin{document}
\maketitle
\begin{abstract}
Given BM25's enduring competitiveness as an information retrieval baseline, we investigate to what extent it can be even further improved by augmenting and re-weighting its sparse query-vector representation. We propose an approach to learning an augmentation and a re-weighting end-to-end, and we find that our approach improves performance over BM25 while retaining its speed. We furthermore find that the learned augmentations and re-weightings transfer well to unseen datasets.
\end{abstract}

\section{Introduction}
Despite the enormous progress in neural information retrieval (IR) techniques over the last several years~\citep[\textit{inter alia}]{karpukhin-etal-2020-dense,xiong2020approximate,dai2019context}, simply using BM25~\citep{robertson1995okapi,crestani1998document} remains a strong baseline approach~\citep{thakur2021beir,sciavolino-etal-2021-simple}. This is especially the case if it is important that retrieval be fast, or that the retrieval method not consume too much memory or disk space. Given the effectiveness of this rather simple baseline, it is natural to wonder whether BM25 might become an even \textit{more} competitive baseline with a bit of additional engineering.

One straightforward approach to improving BM25 while retaining its favorable properties is to do query augmentation --- that is, to augment the query with additional tokens, thereby improving the quality of the retrieved documents. Notably, \citet{nogueira-cho-2017-task} propose to learn query augmentations using reinforcement learning (RL) techniques, with recall-at-$K$ as the reward. While query augmentations are indeed discrete (consisting of word tokens), and therefore certainly congenial to RL methods, we show that augmentations can be much more simply learned end-to-end, in the course of minimizing a standard contrastive loss.

We find that our augmentation approach improves performance, while allowing for retrieval that is as fast, and sometimes even faster, than BM25. Moreover, our approach can be realized by simply modifying the query and IDF vectors (respectively), which allows for doing retrieval with standard tools, such as Pyserini~\citep{lin2021pyserini}. In particular, unlike many recent sparse approaches to retrieval (e.g., SPLADE~\citep{formal2021splade}) we do not need to re-encode the documents. 

Finally, we find that our approach transfers well between datasets. This is an encouraging finding, since transfer can be challenging for neural IR methods, while it is typically not a challenge for BM25. Code for reproducing all models and experiments is available at \url{https://github.com/chenyn66/bm25_aug}.

\section{Augmenting BM25}
Recall that BM25~\citep{robertson1995okapi,crestani1998document} scores the compatibility between a query $q$, viewed as a set of tokens, and a document $d$, also a set. Given a collection $\mcD$ of $N$ documents, and assuming a vocabulary $\mcV$, let $\boldv \in \reals^{|\mcV|}$ be the document collection's inverse document-frequency (IDF) vector, defined as
\begin{align*}
v_i = \log (\frac{N - |\{d \in \mcD \mid w_i \in d\}| + 0.5}{|\{d \in \mcD \mid w_i \in d\}| + 0.5} + 1),
\end{align*}
where $|\{d \in \mcD \mid w_i \in d\}|$ counts the number of documents in the collection in which word type $w_i$ appears. Also let $\boldf(d) \in \reals^{|\mcV|}$ be a document $d$'s BM25 term-frequency vector, defined as
\begin{align*}
\mathrm{f}(d)_i = \frac{\mathrm{count}(w_i, d) (k+1)}{\mathrm{count}(w_i, d) + k(1-b+ \frac{b|d|}{M})},
\end{align*}
where $\mathrm{count}(w_i, d)$ counts the number of times word type $w_i$ appears in document $d$, $k$ and $b$ are hyperparameters, and $M$ is the average length of the documents in the collection. Finally, let $\mathbf{bow}(q) \in \{0,1\}^{|\mcV|}$ be the binary bag-of-words representation of $q$, where $\mathrm{bow}(q)_i$ is 1 if $w_i \in q$, and is 0 otherwise.
The BM25 score between $q$ and $d$ can then be written as
\begin{align*}
\mathrm{BM25}(q, d) = (\boldv \odot \mathbf{bow}(q))^\top \boldf(d),
\end{align*}
where $\odot$ is element-wise multiplication.

\paragraph{Augmenting differentiably}
To augment BM25's query vector, one can produce an additional set of tokens $\hat{q}$, and then score documents with $(\boldv \odot \mathbf{bow}(q \cup \hat{q}))^\top \boldf(d)$. While $\hat{q}$ is of course discrete, we observe that if we are willing to rescale $\boldv$ element-wise by some additional vector $\boldc \in \reals^{|\mcV|}$, we have
\begin{align*}
    \boldc \odot \boldv \odot \mathbf{bow}(q \cup \hat{q}) = \boldv \odot (\mathbf{bow}(q) + \bolda)
\end{align*}
for $\bolda \in \reals^{|\mcV|}$, so long as $a_i = c_i\mathrm{bow}(q \cup \hat{q})_i - \mathrm{bow}(q)_i$. Moreover, we have that $a_i \neq 0$ only if $\mathrm{bow}(q \cup \hat{q})_i = 1$. Thus, we can predict $\bolda$ as our \textit{non-discrete} augmentation vector, and score documents with 
\begin{align*}
(\boldv \odot (\mathbf{bow}(q) + \bolda))^\top \boldf(d),
\end{align*}
and this will be equivalent to scoring with $(\boldc \odot \boldv \odot \mathbf{bow}(q \cup \hat{q}))^\top \boldf(d)$ for some $\boldc$. 

In practice, we use a model to produce both the $\bolda$ vector as well as an additional element-wise weighting vector $\boldw$, as parameterized functions of the query $q$. This leads to the following final scoring function:
\begin{align} \label{eq:score}
\mathrm{score}&(q, d) = \\
&(\boldw(q) \odot \boldv \odot (\mathbf{bow}(q) + \bolda(q)))^\top \boldf(d), \nonumber
\end{align}
where we have written $\boldw(q)$ and $\bolda(q)$ to emphasize that these are (parameterized) functions of the query. This scoring function is differentiable with respect to both $\bolda(q)$ and $\boldw(q)$, and we can therefore train the models producing these vectors with a standard objective, described below. We also describe below how to regularize $\bolda(q)$ to ensure it is sparse.

\paragraph{Retrieval} Since $a_i$ is nonzero only when $q$ or $\hat{q}$ contains $w_i$, at retrieval time we can extract the augmented set $q \cup \hat{q}$ from $\bolda$ to be our augmented query. If we then define a new IDF vector $\boldv'$ with elements $v'_i = w_i v_i (\mathrm{bow}(q)_i + a_i)$, we can use standard BM25 implementations to retrieve the highest-scoring documents using $\boldv'$ as the IDF vector, and this will be equivalent to retrieving documents under $\boldc \odot \boldv \odot \mathbf{bow}(q \cup \hat{q})$ for some $\boldc$.\footnote{Some implementations of BM25, such as Pyserini's, allow rescaling IDF values directly, which is what we do.} 

\paragraph{Parameterization}
We feed a linearized $q$ into a pretrained BERT-like~\citep{devlin-etal-2019-bert} encoder, after first prepending to it a [CLS] token. Let $\enc(q)_0 \in \reals^E$ be the encoder's representation of the [CLS] token, and let $\enc(q)_i \in \reals^E$ be the encoder's representation of the $i$-th token in the query, for $i \geq 1$. We then parameterize $\bolda$ and $\boldw$ as follows:
\begin{align*}
\bolda(q) &= \RELU(\boldW \, \enc(q)_0) \\
w(q)_i &= \RELU(\boldu^\top \enc(q)_i),
\end{align*}
where $\boldW \in \reals^{|\mcV| \times E}$ and $\boldu \in \reals^E$. 

\paragraph{Training} We train end-to-end, using a standard contrastive loss:
\begin{align*}
    \mcL_{rank} = -\log \frac{\exp{(\mathrm{score}(q,d^+))}}{\sum_{d' \in \mcD^- \cup \{d^+\}} \exp{(\mathrm{score}(q,d'))}},
\end{align*}
where $d^+$ is a positive document provided by the training dataset, and $\mcD^-$ consists of hard negatives mined by BM25 as well as in-batch negatives, following the approach of \citet{karpukhin-etal-2020-dense}.

\paragraph{Sparse regularization} Note that in practice, BM25's retrieval speed depends on how many distinct word types appear in the query. In order to promote retrieval efficiency, we regularize $\bolda(q)$ to ensure it is sparse. We found it beneficial to encourage sparsity \textit{more} for frequent words, which are less discriminative. We therefore weight the $L1$ regularization per word by a monotonic function of its document frequency. In particular, we use: 
\begin{align*}
    \mcL_{reg} = \mathrm{sqrt}(\boldh)^\top \bolda(q),
\end{align*}
where $h_i = \frac{|\{d \in \mcD \mid w_i \in d\}|}{N}$, and where the square-root is applied elementwise. Our final training loss is then $\mcL = \mcL_{rank} + \lambda \mcL_{reg}$.

\section{Experiments}
\paragraph{Datasets}
We evaluate the retrieval performance of our proposed method on Natural Questions~\citep[NQ;][]{nq2019}, EntityQuestions~\citep{sciavolino-etal-2021-simple}, and MSMARCO passage ranking task~\citep{bajaj2016ms}. We use TriviaQA~\citep{joshi-etal-2017-triviaqa} and EntityQuestions to test out-of-distribution retrieval.

\paragraph{Experimental Details}
We initialized all models with distilbert-base-uncased~\cite{Sanh2019DistilBERTAD}, using the Hugging Face implementation~\citep{wolf-etal-2020-transformers}. On NQ, models were trained with the AdamW optimizer~\citep{kingma2015adam,loshchilov2018decoupled}, using a learning rate of $3e^{-4}$, a batch size of 144, and a value for $\lambda$ of $0.1$. We trained the models for 45 epochs with 1 hard negative per sample. We used the same settings for training on EntityQuestions but only trained for 10 epochs. On MSMARCO, we used $3e^{-5}$ for the learning rate, 144 for the batch size, and $0.025$ for $\lambda$. Models were trained for 2 epochs with 4 hard negatives per sample. We trained and evaluated our models on a single A6000 GPU. Training took about 70 minutes on NQ and 30 minutes on MSMARCO. Documents were tokenized by BERT's WordPiece tokenizer and indexed by Pyserini~\citep{lin2021pyserini}. We also used Pyserini for retrieval at test time. 

\paragraph{Tokenization} We emphasize that the performance of standard BM25 depends on the tokenization used. Since we use a distilbert model, we must tokenize queries and documents with a WordPiece tokenizer~\citep{kudo-2018-subword,devlin-etal-2019-bert}. Because this is not the default tokenization employed by Pyserini, we report baseline BM25 numbers using both tokenizations. We refer to BM25 with the default Pyserini tokenization as ``BM25 (Pyserini)'' and BM25 with the WordPiece tokenization as ``BM25 (Ours).''

\subsection{Retrieval Performance}

\begin{table}[t!]
\centering
\small
\begin{tabular}{lccc}
\toprule
 \textbf{NQ} & Acc@5 & Acc@20 & Latency \\
 \midrule
BM25 (Pyserini) & 0.436 & 0.629 & 0.099s \\
BM25 (Ours) & 0.430 & 0.589 & 0.103s \\
GAR+BM25 & 0.609 & 0.744 & 5min \\
DPR & 0.668 & 0.781 & 30min\\
SEAL & 0.613 & 0.762 & 35min \\
Ours & 0.557 & 0.694 & 0.146s \\
\midrule
\textbf{EntityQuestions} & Acc@5 & Acc@20 & Latency \\
\midrule
BM25 (Pyserini) & 0.616 & 0.720 & 0.060s \\
BM25 (Ours) & 0.526 & 0.637 & 0.094s \\
DPR & - & 0.684 & - \\
Ours & 0.693 & 0.798 & 0.669s \\
\midrule
\textbf{MSMARCO} & NDCG@10 & R@100 & Latency \\
\midrule
BM25 (Pyserini) & 0.228 & 0.658 & 0.020s \\
BM25 (Ours) & 0.217 & 0.623 & 0.031s \\
SPLADE & 0.433 & - & 1.764s \\
Ours & 0.251 & 0.687 & 0.030s \\
\bottomrule
\end{tabular}
\caption{Retrieval results on NQ, EntityQuestions, and MSMARCO. Non-BM25 baselines include GAR+BM25~\citep{mao-etal-2021-generation}, DPR~\citep{karpukhin-etal-2020-dense}, and SEAL~\citep{bevilacqua2022autoregressive} on NQ, DPR on EntityQuestions, and SPLADE~\citep{formal2021splade} on MSMARCO. Results and latencies for non-BM25 methods are taken from their respective papers. Latency for DPR on NQ is reported by \citet{mao-etal-2021-generation}; the latency number is unavailable for DPR on EntityQuestions. Latency for SPLADE is measured with the Pyserini implementation.}
\label{tab:nq}
\end{table}

We report results and latencies on NQ, EntityQuestions, and MSMARCO in Table~\ref{tab:nq}. We measure per-query latency using wall-clock time on the same machine. On NQ, our method improves 12.1 percentage points in top-5 retrieval accuracy over the vanilla BM25 while adding only 43 milliseconds in latency. Compared to GAR~\citep{mao-etal-2021-generation}, an alternative query augmentation method that autoregressively predicts the target document given a query, our method retrieves much more quickly and performs only slightly worse.

On MSMARCO and EntityQuestions, our method consistently improves over the baseline BM25. Notably, our method achieves lower latency than BM25 on MSMARCO since our weighting allows skipping some terms in the query by setting corresponding $w(q)_i$ to 0. On EntityQuestions, a dataset designed to demonstrate the inconsistency of dense retrievers, our method outperforms both BM25 and DPR. 

\begin{table*}[t!]
\centering
\small
\begin{tabular}{lcccc}
\toprule
 & Accuracy@5 & Accuracy@20 & Latency & Augmentation Length \\
 \midrule
Full Setting & 0.557 & 0.694 & 0.146 & 12.334 \\
- w/o Weighted $L1$ & 0.562 & 0.704 & 0.268 & 15.211 \\
- w/o Weight & 0.545 & 0.683 & 0.269 & 19.165 \\
- w/o BM25 Scoring & 0.487 & 0.635 & 0.225 & 31.861 \\
- w/o BM25 Scoring \& Weighted $L1$ & 0.525 & 0.670 & 0.377 & 21.794 \\
\bottomrule
\end{tabular}
\caption{Ablation experiments on NQ. From top to bottom, we consider our approach with a uniform weighting of the $L1$ penalty (rather than by word frequency), without the elementwise weight vector $\boldw$, using just a bag-of-words rather than BM25-style query and document representations, and with both uniform $L1$ and bag-of-words representations.}
\label{tab:ablation}
\end{table*}

\subsection{Transfer Results}
\label{sec:transfer}
\begin{table}[t!]
\centering
\small
\begin{tabular}{lcc}
\toprule
 & Accuracy@5 & Accuracy@20 \\
 \midrule
\textbf{TriviaQA} & &  \\
\midrule
BM25 (Pyserini) & 0.677 & 0.773 \\
BM25 (Ours) & 0.636 & 0.742 \\
Ours & 0.662 & 0.755 \\
\midrule
\textbf{EntityQuestions} &  &  \\
\midrule
BM25 (Pyserini)& 0.616 & 0.720 \\
BM25 (Ours)& 0.526 & 0.637 \\
Ours & 0.542 & 0.656 \\
\bottomrule
\end{tabular}
\caption{Results of our approach trained on NQ and then transferred to TriviaQA (top) and EntityQuestions (bottom).}
\label{tab:transfer}
\end{table}
One of the major concerns relating to language-model-assisted retrieval is that it may not generalize well out-of-distribution. To test our method in a transfer setting, we select the best model trained on NQ, and test it on TriviaQA and EntityQuestions without further fine-tuning. We show the results of this experiment in Table~\ref{tab:transfer}, where we see that our method generalizes to both datasets, and improves from the baseline by 2-3 percentage points. At the same time, it is clear from comparing the results of BM25 (Pyserini) with BM25 (Ours) that the Pyserini tokenization is helpful for these datasets. We anticipate being able to further improve given a pretrained model using the preferred tokenization, which we leave to future work.

\subsection{Ablation Study}

In Table~\ref{tab:ablation} we present the results of ablating various aspects of our approach on the NQ dataset. Our full setting achieves the best balance between accuracy and efficiency. Compared to a uniform $L1$ penalty, our $L1$ penalty weighted by document frequency achieves lower latency while predicting augmentations of similar lengths. This suggests that the model is augmenting with rarer terms, thereby reducing the total number of documents in the inverted index and increasing speed. We also see that the additional weighting $\boldw$ is helpful. Finally, we check whether using BM25-style query and document vectors, rather than simple bag-of-words vectors, is important, by replacing the scoring function~\eqref{eq:score} with the following: 
\begin{align*}
(\boldw(q) \odot (\mathbf{bow}(q) + \bolda(q)))^\top \mathbf{bow}(d).
\end{align*} 
We observe that this also decreases performance.

\section{Related Work}
While recent dense retrievers have shown strong retrieval performance~\citep{reimers-gurevych-2019-sentence,karpukhin-etal-2020-dense}, their high latencies limit their application to first-stage retrieval. To improve efficiency, late-interaction approaches have been proposed \citep{khattab2020colbert,gao-etal-2021-coil,formal2021splade}. Here, documents are first retrieved with an inverted-index and then scored by aggregating term embeddings pre-computed while indexing. Although such methods reduce retrieval latencies, the need to store dense representations of documents significantly increases index sizes~\citep{thakur2021beir}.

Document expansion methods, such as Doc2query~\citep{nogueira2019document, nogueira2019from}, allow indexing and retrieval using standard BM25. \citet{nogueira2019from} use a language model to predict possible queries given a document; augmented documents are then constructed by appending these queries. By adding to the document, this approach addresses the term mismatching issue affecting sparse retrieval methods. However, this approach also requires running a language model over every document, which is expensive, especially if new documents are added incrementally. It is also potentially infeasible to do this when documents are very long. In contrast, we restrict our method to only perform neural operations on the queries, which are usually much shorter than documents. 

To the best of our knowledge, only a few recent methods meet this requirement of only modifying queries. \citet{nogueira-cho-2017-task} use reinforcement learning to predict discrete augmentations. GAR~\citep{mao-etal-2021-generation} and SEAL~\citep{bevilacqua2022autoregressive} train language models to generate target documents or n-grams. Our approach differs from these methods by optimizing for BM25 retrieval in an end-to-end fashion, and in being significantly faster.

\section{Conclusion}
We propose a novel approach for learning to augment BM25 end-to-end with a language model. Our method improves over BM25 on three different datasets while retaining its efficiency. Additionally, we show that such improvements are able to generalize out-of-distribution. With its simple formulation, our method can be easily integrated into existing sparse retrieval frameworks. And we believe it might serve as a stronger sparse baseline for future work in retrieval.

\section*{Limitations}
As mentioned in Sec.~\ref{sec:transfer}, tokenization methods heavily influence retrieval performance. This is a limitation both of BM25 and of our modification of it. In its current form, there are no straightforward solutions that allow our method to augment queries with words rather than the subword tokens of the pretrained tokenizer. Thus, in situations where word-tokenization is important for BM25 (some of which appear in Table~\ref{tab:nq}), using our method would require pre-training a word- rather than subword-based model, which may be difficult.

Relatedly, since our method makes fundamental use of a pretrained model as the backbone, it suffers from the same problems afflicting pretrained models, including susceptibility to misinformation and bias, and requiring significant computational resources.

\section*{Ethics Statement}
As our approach attempts to improve retrieval technology, the ethical considerations are similar to those of other retrieval technologies, especially those utilizing large pretrained language models. In particular, there is always a risk that the documents retrieved by our approach will be influenced by errors or biases in the underlying model, and it is necessary to ensure this does not happen before deployment. Because our augmentations are token-based, rather than based on dense representations, it should be slightly easier to manually check whether augmentations are problematic. We also emphasize that our approach is relatively undemanding computationally, which we believe to be a positive feature.

\bibliography{anthology,custom,zotero_xc}
\bibliographystyle{acl_natbib}

\appendix
\section{Dataset Statistics}
\begin{table}[h!]
\centering
\small
\begin{tabular}{lccc}
\toprule
 Dataset & Train & Dev & Test \\
 \midrule
Natural Questions & 58,880 & 8,757 & 3,610  \\
EntityQuestions & 176,560 & 22,068 & 22,075 \\
TriviaQA & 60,413 & 8,837 & 11,313 \\
MSMARCO & 502,939 & 6,980 & -\\
\bottomrule
\end{tabular}
\caption{Number of train/dev/test queries on each dataset. On MSMARCO, we follow the same approach as the previous work that reports the dev set performance of BEIR \citep{thakur2021beir}.}
\label{tab:dataset}
\end{table}
\section{License}
All packages and datasets used in our study are released with Apache-2.0 or MIT licenses.
\label{sec:license}

\end{document}